\documentclass{article}

\usepackage{arxiv}

\usepackage[utf8]{inputenc} 
\usepackage[T1]{fontenc}    
\usepackage{hyperref}       
\usepackage{url}            
\usepackage{amsfonts}       
\usepackage{nicefrac}       
\usepackage{microtype}      
\usepackage{multirow}
\usepackage{tablefootnote}
\usepackage{footnote}
\usepackage{graphicx}
\usepackage{afterpage}
\usepackage[numbers,sort&compress]{natbib}
\usepackage{fullpage}
\usepackage{caption}
\usepackage{needspace}
\usepackage{times}
\usepackage{fancyhdr,graphicx,amsmath,amssymb}
\usepackage[linesnumbered,ruled]{algorithm2e}

\title{Applications of shapelet transform to time series classification of earthquake, wind and wave data}

\author{
  Monica Arul\\
  NatHaz Modeling Laboratory\\
  Department of Civil Engineering\\
  University of Notre Dame\\
  Notre Dame, IN 46556 \\
  \texttt{maruljay@nd.edu} \\
   \And
 Ahsan Kareem \\
  NatHaz Modeling Laboratory\\
  Department of Civil Engineering\\
 University of Notre Dame \\
  Notre Dame, IN 46556 \\
  \texttt{kareem@nd.edu} \\
}

\begin{document}
\maketitle

\begin{abstract}
Autonomous detection of desired events from large databases using time series classification is becoming increasingly important in civil engineering as a result of continued long-term health monitoring of a large number of engineering structures encompassing buildings, bridges, towers, and offshore platforms. In this context, this paper proposes the application of a relatively new time series representation named “Shapelet transform”, which is based on local similarity in the shape of the time series subsequences. In consideration of the individual attributes distinctive to time series signals in earthquake, wind and ocean engineering, the application of this transform yields a new shape-based feature representation. Combining this shape-based representation with a standard machine learning algorithm, a truly “white-box” machine learning model is proposed with understandable features and a transparent algorithm. This model automates event detection without the intervention of domain practitioners, yielding a practical event detection procedure. The efficacy of this proposed shapelet transform-based autonomous detection procedure is demonstrated by examples, to identify known and unknown earthquake events from continuously recorded ground-motion measurements, to detect pulses in the velocity time history of ground motions to distinguish between near-field and far-field ground motions, to identify thunderstorms from continuous wind speed measurements, to detect large-amplitude wind-induced vibrations from the bridge monitoring data, and to identify plunging breaking waves that have a significant impact on offshore structures.  
\end{abstract}


\section{Introduction}
With the wider availability of sensor technology through easily affordable sensor devices, a number of fields now use sensors which have led to a deluge of high-dimensional time series data collected continuously over time. Given the pervasiveness of time series data, there has been a growing interest in using data mining and machine learning techniques for time series analysis. While there are many different time series mining techniques, they all have one feature in common: high-level representation of the raw data. Several time series representations have been discussed in the literature such as Fourier transforms, wavelet transform \cite{gurley1999applications}, piecewise polynomials, symbolic mappings \cite{alves2015structural}, singular value decomposition \cite{magalhaes2008dynamic} and so on. However, most of these methods hold little utility for time series data mining tasks such as indexing, anomaly detection, classification, clustering, prediction, etc. Moreover, many representation techniques abandon the precise values of the measured time series for high-level approximations and thus do not conserve the general shape of the original time series. This can reduce comprehensibility and make straight forward visual interpretation challenging especially for non-specialists. 

This paper proposes the use of “Shapelets”, a shape-based approach for time series representation, for engineering applications, which emphasizes the fundamental shape characteristics of time series. Time series shapelets stem from the desire to emulate human’s innate capacity to visualize the shape of data and identify almost instantly similarities and differences between patterns. Shapelets help computers to perform this complex task by identifying the local or global similarity of shape that can offer an intuitively comprehensible way of understanding long time series. The concept of shapelets was first presented by \cite{ye2009time} as phase-independent time-series sub-sequences that are highly discriminative and informative. The shapelets are used to transform data into a local-shape space where each feature is the distance between a shapelet and a time series \cite{lines2012shapelet}. The result of this transform is that the new representation can be applied to any standard machine learning algorithm, facilitating analysis based on local shape-based similarity. This renders the method as a complete “white-box” machine learning model involving understandable and easily visualizable features that makes the entire process transparent and completely open to inspection. This, in turn, increases the interpretability of the model compared to the other state-of-the-art black-box machine learning models and helps domain practitioners gain better insights from their data. 

Moreover, the effectiveness of all machine learning models is dependent on the choice of features used to characterize the data. In most of the cases, it is intractable even for the domain experts to know the best features to extract a priori. Shapelet transform, can help circumvent the manual feature construction process by providing a universal standard feature for all types of engineering data independent of the process characteristics unique to them. Besides, the universal feature extraction capability of the shapelet transform can help to truly automate the event detection process without the intervention of domain practitioners. It is envisaged that this approach will flourish in the near future as more time series databases get populated and the need for autonomous detection becomes increasingly desirable.

Ever since their introduction, shapelets have been used in various fields for time-series mining tasks such as characterization, clustering, classification and anomaly detection \cite{mueen2011logical,zakaria2012clustering,grabocka2014learning,hills2014classification,beggel2019time,ye2011time}. In this paper, we present numerous examples where we have found shapelets useful in the studies involving earthquake, wind and wave data. The shapelet transform method is applied to the automated detection of earthquake events from continuously recorded ground-motion measurements. Unlike several similarity-based seismological approaches that are restricted to detecting just the replicas of previously recorded events, shapelets were able to detect both cataloged and a plethora of unknown earthquake events thus helping to significantly reduce the catalog completeness magnitudes. Also, shapelet transform is applied to identify strong pulses in the velocity time history of ground motions thus helping to clearly distinguish between near-field and far-field ground motions.

The second application of the shapelet transform is to wind and wind-induced vibration data. Shapelet transform is used to identify thunderstorms from continuous wind speed measurements without the help of meteorological parameters or expert judgment thus making it a suitable tool for automatic extraction and cataloging of thunderstorms from long-term wind monitoring networks. Shapelet transform is also applied to wind-induced bridge vibration data to identify large-amplitude vortex-induced vibrations (VIV) from noise and other random excitations like buffeting and vehicle-induced vibrations. This capability of shapelets can help long-term bridge health monitoring networks to identify and monitor these critical vibrations that are known to cause fatigue failure of components and discomfort to drivers. Finally, the shapelet transform is applied to detect plunging breaking waves that form a most significant type of load on offshore structures.

\section{A brief overview of shapelet transform}

Consider time series 1 and 2 generated as a result of an event as shown in Fig. 1. Both time series have long stretches of aperiodic waveforms. However, a local shape appears for a short duration that differs substantially from the rest of the time series. These localized shapes are called shapelets. These discriminatory shapes which are phase and amplitude-independent serve as a powerful feature for time series mining tasks. Shapelets are based on a local shape-based approach for analyzing and classifying time series that focuses on highly informative subsequences of time series. Analysis methods based on the global attributes of time series are unintuitive and reduce comprehensibility. By examining local-shape-based features, we ensure that these small discriminatory shapes are not averaged out but rather used to distinguish the time series, exactly as they are under intuitive visual inspection. One could draw a similarity with the wavelet transform as it is localized and preserves local features while in global schemes like the Fourier transform, local features are averaged. Shapelets are not only helpful for time series classification but can also enhance understanding of time-series data for domain experts. Classification of time series based on shapelets uses a similarity measure, in the present case Euclidean distance, between the time series and the shapelet as a discriminatory feature to classify time series. Distance calculations decide the presence or absence of a shapelet in a particular time series. Each subsequence in each time series is considered as a potential shapelet candidate. Thus, shapelets are discovered through an extensive search for every possible shapelet candidate of all possible lengths in every time series. Shapelet transform has five major stages: generation of shapelet candidates, distance calculation between a shapelet and a time series, assessment of the quality of shapelets, discovery of shapelets, and data transformation.

\begin{figure}[htb]
  \centering
  \captionsetup{justification=centering}
  \includegraphics[scale=0.7]{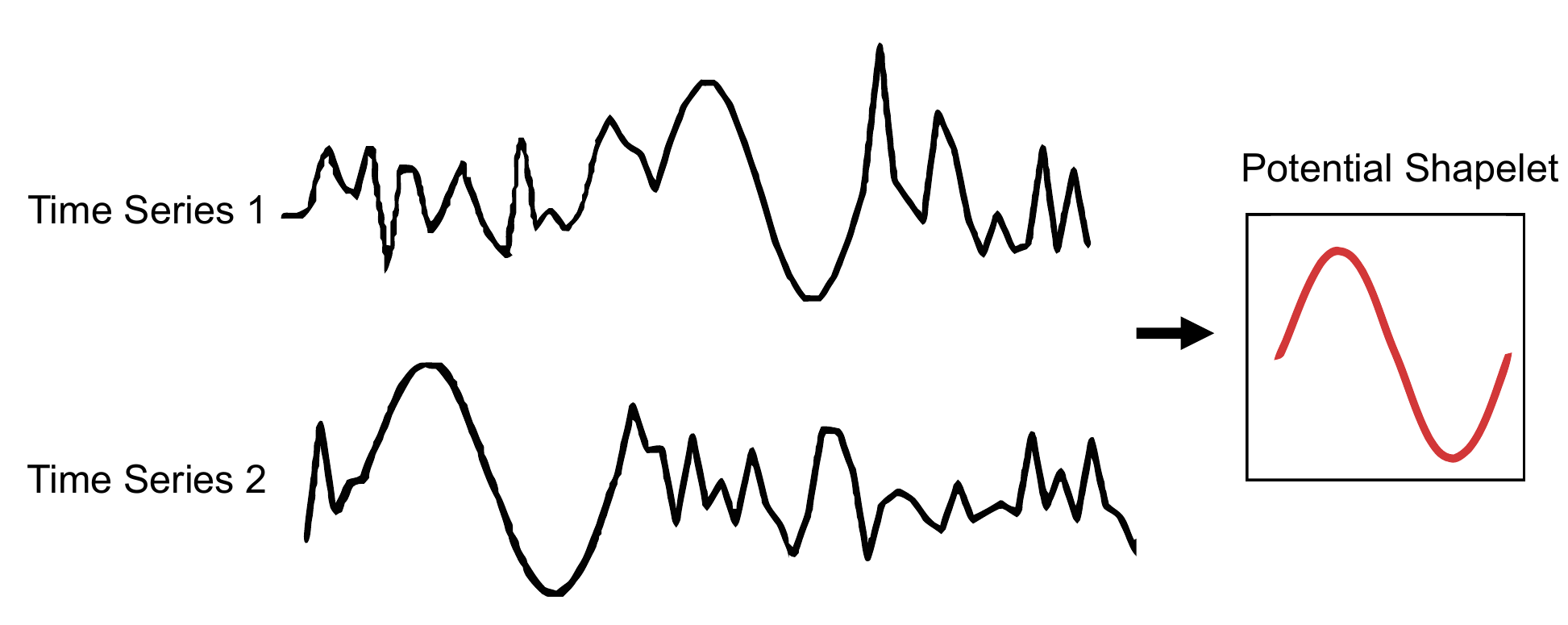}
  \caption{Time series shapelets}
  \label{fig:fig1}
\end{figure}

\begin{figure}[htbp]
  \centering
  \captionsetup{justification=centering}
  \includegraphics[scale=0.85]{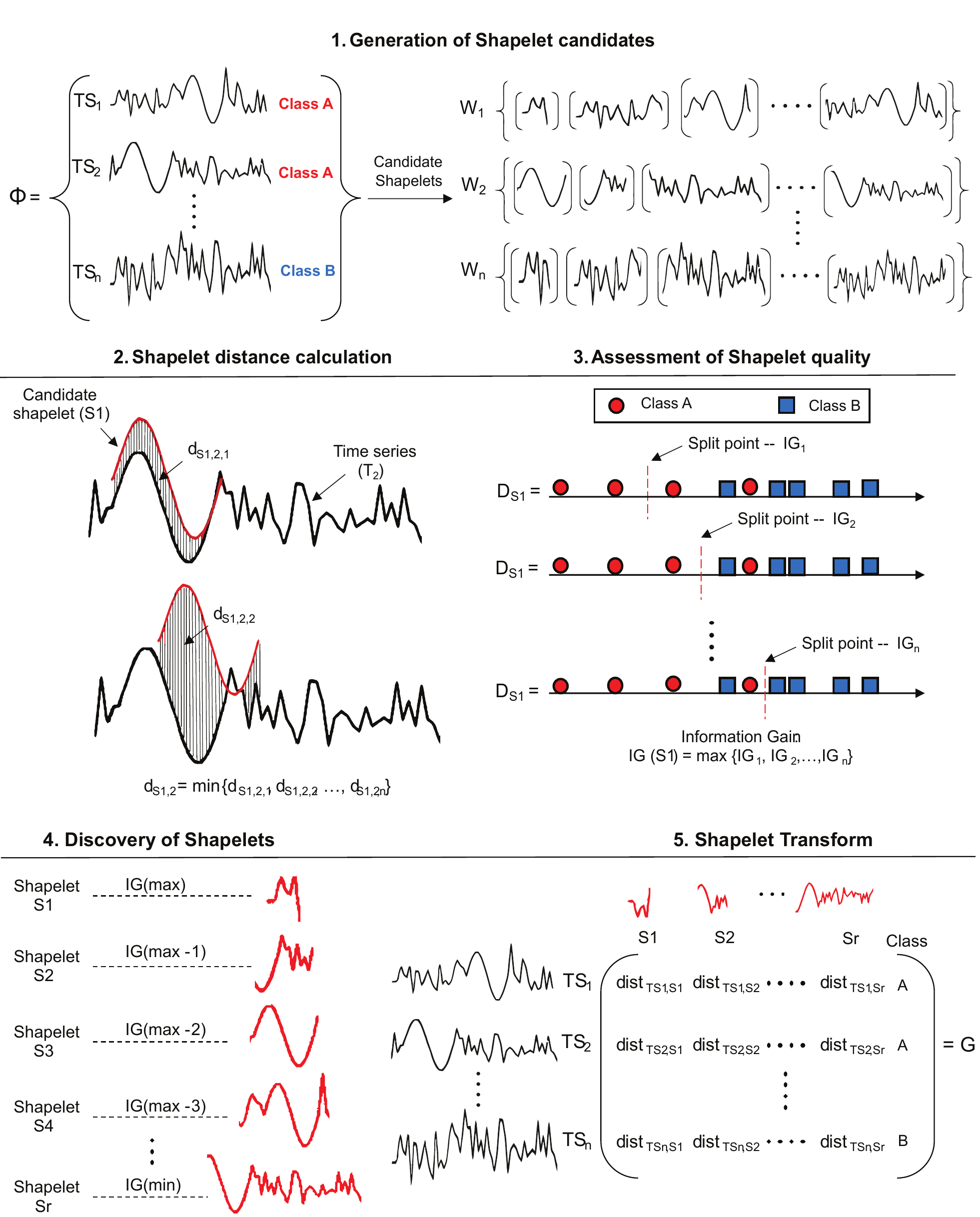}
  \caption{Overview of shapelet transform}
  \label{fig:fig2}
\end{figure}

\subsection{Generation of shapelet candidates}
Consider a time-series dataset $TS=\left\{ {{TS}_{1}},{{TS}_{2,}}.....,{{TS}_{n}} \right\}$ where an individual time series ${{TS}_{i}}=\left\langle {{t}_{i,1}},{{t}_{i,2}},...,{{t}_{i,m}} \right\rangle $ is an ordered set of \textit{m} real numbers. Let \textit{C} be the set of corresponding class labels for each time series. A time series learning set $\Phi \left\{ TS,C \right\}$ is first created by a vector of instance input-output pairs ${{\Phi }_{i}}=({{TS}_{i}},{{C}_{i}})$ as shown in step 1 of Fig. 2.  Each subsequence in each time series in ${\Phi}$ is considered as a potential shapelet candidate. So, there are $\left( m-l \right)+1$ discrete subsequences of length \textit{l} between a subsequence \textit{X} of length \textit{l} of a time series \textit{TS} of length \textit{m}. If ${{W}_{1}}$ is the set of all candidate shapelets of length \textit{l} in a time series ${{TS}_{1}}$, then 

\begin{equation}
	\ {W}_{1}=\left\{ {{w}_{\min }},{{w}_{\min +1}},...,{{w}_{\max }} \right\}\
\end{equation}

where $min\ge 3$  as it is the minimum meaningful length for a time series and  $max\le m$.

\subsection{Shapelet distance calculation}
Euclidean distance is used as a similarity measure in shapelets and the squared Euclidean distance between a subsequence \textit{X} of length \textit{l} and another subsequence \textit{Y} of the same length is defined as:

\begin{equation}
	\ dist(X,Y)=\sum\limits_{i=1}^{l}{{{\left( {{x}_{i}}-{{y}_{i}} \right)}^{2}}}\
\end{equation}

As shown in step 2 of Fig.2, the distance between a potential shapelet candidate and all series in \textit{TS} is computed to create a list of \textit{n} distances called an orderline ${D}_{S}$. An orderline consists of distance values and the class label corresponding to the time series for which the distance value is calculated. The orderline is then sorted in increasing order of the distance value. Thus, the distance between a shapelet candidate ${S}_{1}$ and all time series in \textit{TS} is given by,

\begin{equation}
    	\ {{D}_{S}}=\left\langle {{d}_{S1,1}},{{d}_{S1,2}},...,{{d}_{S1,n}} \right\rangle \
\end{equation}

It is a time-consuming task to calculate ${D}_{S}$ and hence a number of speed-up techniques have been proposed in the literature to handle the large volume of calculations. \cite{ye2009time,mueen2011logical,ye2011time,hills2014classification,rakthanmanon2013fast}

\subsection{Assessment of shapelet quality}
Information Gain (IG) \cite{shannon1949mathematical} is used as the standard approach to calculate the quality of a shapelet \cite{ye2009time,mueen2011logical,ye2011time}. If a time series dataset \textit{T} can be split into two classes, \textit{X} and \textit{Y}, then the entropy of \textit{T} is: 

\begin{equation}
    \ H(T)=-p(X)\log (p(X))-p(Y)\log (p(Y))\
\end{equation}
	
where \textit{p(X)} and \textit{p(Y)} are the proportion of objects in class \textit{X} and \textit{Y} respectively. Thus every splitting strategy partitions the dataset \textit{T} into two sub-datasets ${T}_{X}$ and ${T}_{Y}$. The Information Gain of this split is the difference between the entropy of the entire dataset and the sum of the weighted average of entropies for each split. In the present case, the splitting rule is based on the distance from the shapelet candidate S to every series in the dataset.  The best possible shapelet will generate small distance values when compared to a time series of its own class and large distance values for time series from the other class. Thus the best arrangement for the orderline is to have all the distance values corresponding to the class of the shapelet in ${T}_{X}$ and the other in ${T}_{Y}$. Thus, the information gain for each split is calculated as:

\begin{equation}
    \ IG=H(T)-\left( \frac{|{{T}_{a}}|}{|T|}H({{T}_{a}})+\frac{|{{T}_{b}}|}{|T|}H({{T}_{b}}) \right)\
\end{equation}

where $0\le IG\le 1$.

For example, consider the first orderline ${D}_{S_1}$ in step 3 of Fig.2. The orderline consists of 9 distances with 4 distances of class A and 5 of class B. The optimal split point is indicated using a dashed line in Fig.2. The order line splits the 9 distances into two sets. The left hand side contains 2 distances of class A and right hand side contains 1 distance of class A and 5 distances of class B. Using Eq. 4 and 5, the information gain of this split can be calculated as

\begin{equation}
    \ I G_{1}=\left[-\frac{4}{9} \log \left(\frac{4}{9}\right)-\frac{5}{9} \log \left(\frac{5}{9}\right)\right]-\left[\left(\frac{2}{9}\left(-\frac{2}{2} \log \left(\frac{2}{2}\right)\right)\right)+\left(\frac{7}{9}\left(-\frac{2}{7} \log \left(\frac{2}{7}\right)-\frac{5}{7} \log \left(\frac{5}{7}\right)\right)\right)\right]\
\end{equation}

This way, the information gain for all possible split points is calculated and the split point with maximum information gain is selected.

\subsection{Discovery of shapelets}
An algorithm combining all of the above mentioned components of shapelet discovery was developed by \cite{bagnall2017great} and is available at \cite{TimeSeri72:online}. The same algorithm has been adopted and modified to suit the datasets under consideration for the present study. Algorithm 1 gives a pseudo-code overview of the process.

\begin{algorithm}
\DontPrintSemicolon
\SetKwData{Left}{left}\SetKwData{This}{this}\SetKwData{Up}{up}
\SetKwFunction{Union}{Union}\SetKwFunction{FindCompress}{FindCompress}
\SetKwInOut{Input}{input}\SetKwInOut{Output}{output}
\Input{TS (time series dataset) 
\\min (min shapelet length, default = 3) 
\\max (max shapelet length, default = length of time series in TS)
\\r (maximum number of shapelets to store, default = 10*size of TS)
\\quality (predefined information gain threshold, default = 0.05)}
\Output{Shapelets}
\BlankLine
$rShapelets \longleftarrow \Phi$\;
$numC \longleftarrow class$ $distribution$ $(TS)$\;
$p \longleftarrow r$$/$$numC$\;
\ForAll{$TS_{i}\in TS$}{
$shapelets \longleftarrow \Phi$\;
\For{$l \longleftarrow min$ $to$ $max$}{
$W_{i,l} \longleftarrow generate$ $shapelet$ $candidates$ $(TS_{i},min,max)$\;
\ForAll{$subsequences S\in  W_{i,l}$}{
$D_{S} \longleftarrow calculate$ $distances$ $(S, W_{i,l})$\;
$quality \longleftarrow evaluate$ $candidate$ $shapelets$ $(S, D_{S})$\;
\ $shapelets.add$ $(S, quality)$
}
}
\ $group$ $by$ $quality$ $(shapelets)$\;
\ $remove$ $similar$ $(shapelets)$\;
\ $Shapelets \longleftarrow merge$ $(p, rShapelets, shapelets)$\;
}
\caption {Discovery of Shapelets}
\end{algorithm}

The input to the algorithm are the individual time series (\textit{TS}) in the training set, minimum (\textit{min}) and maximum (\textit{max}) length of shapelets and the number of shapelets to store (\textit{r}). As mentioned in Section 2.1, the default minimum length of the shapelets is set to 3 and the maximum length is equal to the length of the time series. The number of shapelets to store is set to a default of 10 times the number of time series in the training set \textit{TS}. Moreover, based on the number of classes (\textit{numC}) in the training set, a limit of \textit{r/numC} shapelets for each class is set as the maximum number of shapelets to store per class. For example, for a binary classification problem, if the training set \textit{TS} contains 20 time series, the maximum number of shapelets to store (\textit{r}) is 200 and the maximum shapelets to store per class are 100. This makes sure that a large number of shapelets from one class will not oust shapelets from another class. The attribute quality, which is the minimum information gain threshold, has a default value of 0.05. This makes sure that poor quality shapelets below this threshold are removed during the shapelet finding process. Using the provided parameters, the algorithm then makes a single pass through the time series data in \textit{TS} taking each subsequence of every time series as a potential shapelet candidate. The generated shapelet candidates are also normalized to make them independent of scale and offset. The distance between each shapelet candidate and time series in the training dataset is calculated and the order list ${D}_{S}$ is formed to assess the quality of shapelets using Information Gain. Once all the shapelets in a time series have been assessed, the poor quality shapelets are removed and the rest is added to the shapelet set. After all the time series in the training set have been evaluated this way, the algorithm returns the discovered shapelets. 

\subsection{Shapelet transform}
Shapelet transform \cite{lines2012shapelet,hills2014classification} is the transformation of time series data into a local-shape space where each feature is the distance between a shapelet and a time series. Given a set of a time series dataset \textit{TS} containing \textit{n} time series and a set of \textit{k} discovered shapelets, the shapelet transform algorithm calculates the minimum distance between each discovered shapelet and each time series in the dataset as shown in Algorithm 2. This transformation creates an \textit{n x k} matrix as illustrated in step 5 of Fig.2 where each element is the minimum Euclidean distance between each shapelet and time series, with the class values appended to the end of each row.

\begin{algorithm}
\DontPrintSemicolon
\SetKwData{Left}{left}\SetKwData{This}{this}\SetKwData{Up}{up}
\SetKwFunction{Union}{Union}\SetKwFunction{FindCompress}{FindCompress}
\SetKwInOut{Input}{input}\SetKwInOut{Output}{output}
\Input{TS containing n time series 
\\Shapelets (S) from algorithm 1 containing k shapelets}
\Output{Shapelet Transform (G)}
\BlankLine
$G \longleftarrow \Phi$\;
\ForAll{$TS_{i}\in TS$}{
\ForAll{$S_{j}\in Shapelets$}{
${G}_{ij}$ $=$ $dist$ $(TS_{i}, S_{j})$
}
}

\caption {Shapelet Transform}
\end{algorithm}

\subsection{Time series classification using shapelets}
While shapelets can be applied to both unsupervised \cite{zakaria2012clustering,grabocka2014learning,zakaria2016accelerating,ulanova2015scalable,zhang2016unsupervised} and supervised time series problems, for brevity we will focus only on the applicability of shapelets for classification in this work. The shapelet based classifier originally developed by \cite{ye2009time} embeds shapelet finding in a decision tree classifier where shapelets are found at every node. Many researchers ever since have demonstrated that higher accuracy can be achieved by using shapelets with more complex classifiers or ensemble of classifiers than with decision trees, where overfitting is a major issue.  \cite{lines2012shapelet,hills2014classification,bagnall2017great,bostrom2017shapelet}. For the present study, Random Forest \cite{breiman2001random} is used as a classifier for time series classification. The Random Forest algorithm seeks to solve the issues with decision trees by classifying examples through using a multitude of decision trees and predicting the class of a sample based on the mean probability estimate across all the trees. Thus, the time series classification using shapelets has two stages. Firstly, the trained shapelet-based classifier searches for shapes similar to the discovered shapelets in every time series and then classifies a series as either containing the particular shape or not. Secondly, the class probability for each prediction is estimated based on the mean predicted class probabilities of all the trees in the forest. For example, if a time series is predicted as A, the shapelet-based classifier also returns a probability for that prediction, i.e. prob(A) = 87\% and prob(B) = 13\%. This makes the classification of time series more transparent and interpretable and helps the user make more informed decisions

\section{Applications of shapelet transform}
In terms of applicability, shapelets have been utilized in a wide variety of domains including motion-capture \cite{ye2009time,lines2012shapelet,ye2011time,hartmann2010gesture,shajina2012human}, spectrographs \cite{ye2009time,ye2011time}, tornado prediction \cite{mcgovern2011identifying}, medical and health informatics \cite{ghalwash2013extraction,xing2011extracting,xing2012early} among others.  A variety of examples are provided herein to demonstrate the applications of shapelets to wind, waves, and earthquakes.

\subsection{Applications of shapelets to earthquakes}
\subsubsection{Automated detection of earthquake events}
Various methods have been developed in the past few decades to detect earthquakes from a continuous time series. Short-term average/Long-term average (STA/LTA) \cite{allen1978automatic,withers1998comparison} and similarity-based search \cite{gibbons2006detection,shelly2007non,peng2009migration,kato2012propagation,skoumal2014optimizing,ross2017aftershocks,beauce2017fast,chamberlain2018eqcorrscan} is the most commonly used algorithms. In recent years, there has been growing interest in using Machine Learning (ML) for automated earthquake detection and picking of seismic arrivals from earthquake data \cite{dai1995automatic,wang1995artificial,tiira1999detecting,zhao1999artificial,wiszniowski2014application,perol2018convolutional,ross2018generalized,wu2018deepdetect,yoon2015earthquake,bergen2018detecting}. Though the above approaches have met with some success, they have their own limitations such as failure to detect weak earthquakes, sensitivity to noise, need for prior knowledge of templates and detection of mere repeating earthquake events.  Shapelets can overcome many of the above-mentioned limitations for automated detection and cataloging of earthquakes. Since shapelets are amplitude and phase-independent, their detection sensitivity is irrespective of the magnitude of the earthquake and the time of occurrence. Unlike other similarity-based approaches which are restricted to detecting just the replicas of previously recorded events, shapelets can also detect unknown earthquake events. The detection capability of shapelets is tested on one week of continuous seismic data as used in \cite{yoon2015earthquake} for comparison purposes. 

The continuous earthquake waveform data (8th January 2011,00:00:00 to 15th January 2011,00:00:00) is measured near the Calaveras Fault by the NC network at station CCOB.EHN and is extracted from the Northern California Seismic Network (NCSN). The continuous data is preprocessed by applying a 4- to 10-Hz bandpass filter to remove noise at lower frequencies. The denoised dataset is decimated from its original sampling frequency of 100 Hz to 20 Hz. It is very important to select the right time window to segment the continuous time-histories for mainly two reasons. A brute search for shapelets in n time series of length m has a complexity of   . Hence a large time window will render the method untenable. Secondly, the dataset under consideration is ridden with low-amplitude noise that can be easily mistaken for weak earthquake events. So, a small time window will not provide seismically relevant shapelets.Apart from continuous time histories, NCSN also provides a catalog of 24 earthquake events and aftershocks that happened on the Calaveras Fault between 8 and 15 January 2011 with their time of occurrence, magnitude, and location. Upon close inspection of these cataloged events, a time window of 5 minutes was able to distinctly capture the occurrence of earthquake events and aftershocks in most cases. Hence a time interval of 5 minutes is chosen, and the continuous seismic time series is broken down into 5-minute chunks. Thus, each time history now contains 6000 data points and a total of 2004 time series datasets are obtained this way for the 1-week period under consideration.  It should be noted that the 5-minute time interval may capture more than one seismic event and subsequent analysis may be required to analyze and extract events within this window. 

\begin{figure}[htbp]
  \centering
  \captionsetup{justification=centering}
  \includegraphics[scale=0.80]{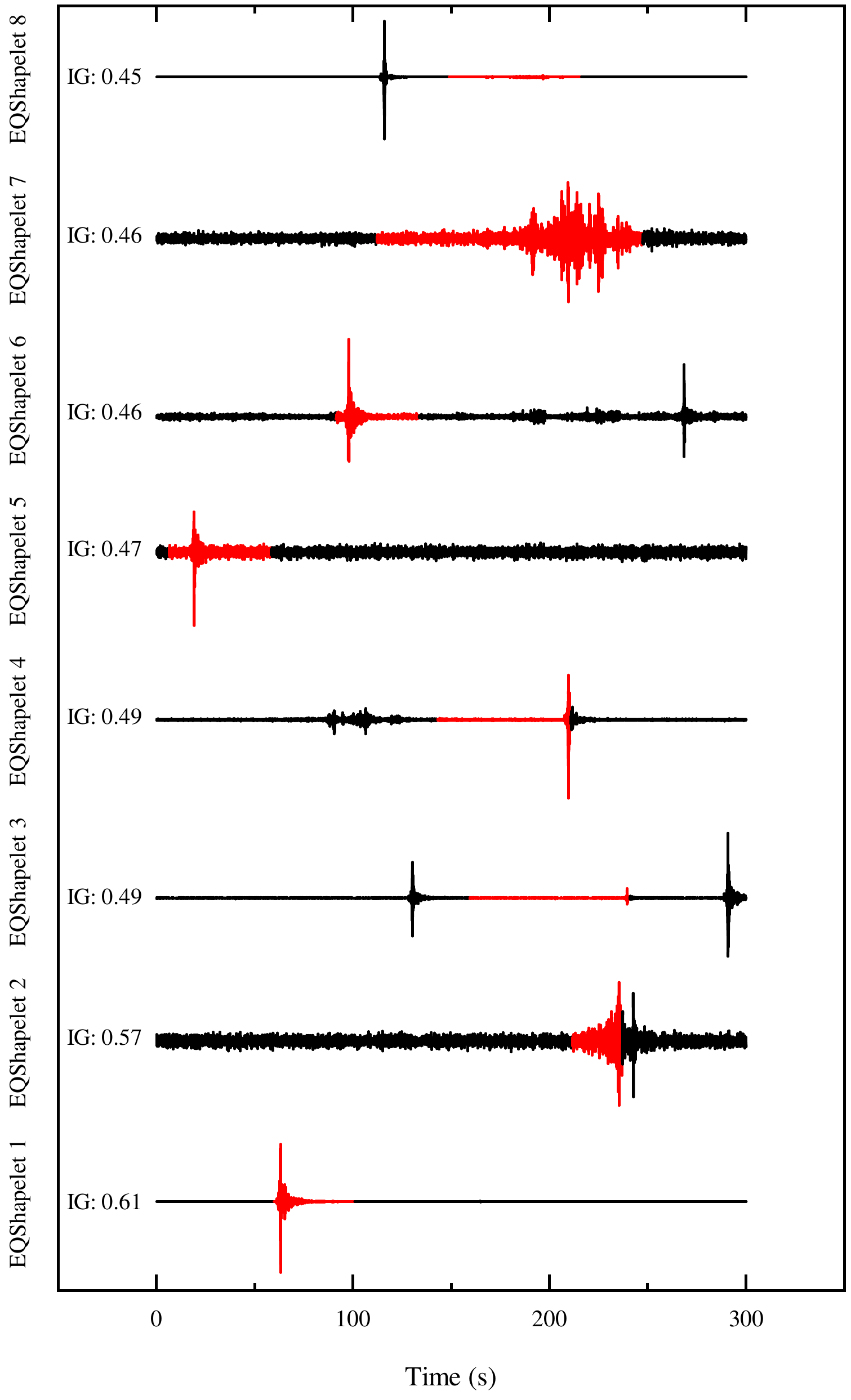}
  \caption{Discovery of EQShapelets from earthquake waveforms recorded on 8th January 2011 at station CCOB.EHN}
  \label{fig:fig3}
\end{figure}

\begin{figure}[htbp]
  \centering
  \captionsetup{justification=centering}
  \includegraphics[scale=0.80]{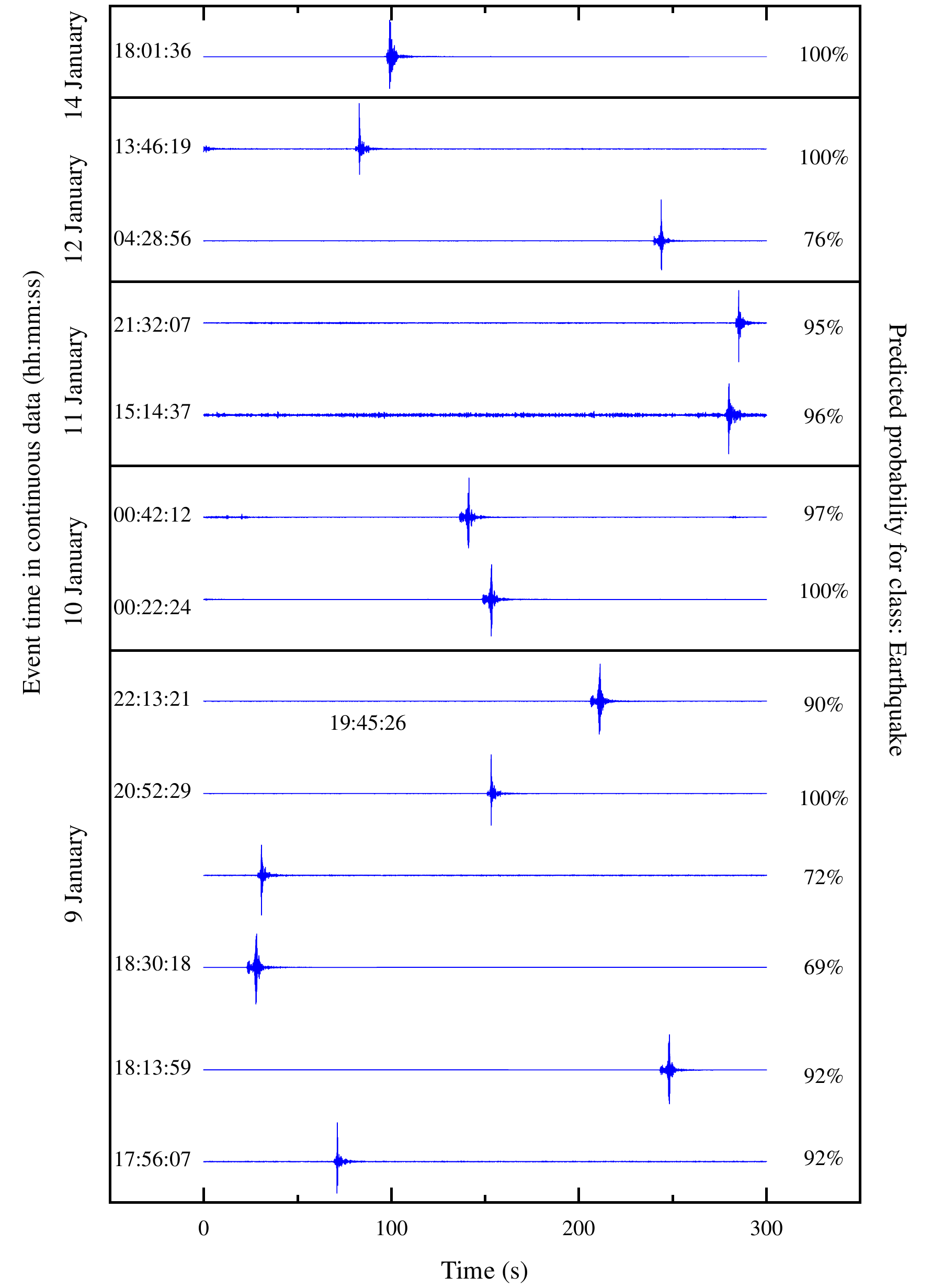}
  \caption{Cataloged earthquake waveforms detected by EQShapelets ordered by event time recorded between 9 and 14 January 2011 from CCOB.EHN}
  \label{fig:fig4}
\end{figure}

According to the NCSN catalog, an Mw 4.1 earthquake occurred on the Calaveras Fault on 8 January 2011, followed by several aftershocks. Thus, the continuous time history recorded between 8 January 2011 (00:00:00) and 9 January 2011 (00:00:00) is used for the discovery of EQShapelets.  A time window of 5 minutes is used to segment the continuous time history into 288 smaller datasets. Out of the 288 datasets, 52 time histories corresponding to earthquake events are manually labeled as “Earthquake events” and 52 other noisy time-history records that do not contain any earthquake events are selected and labeled as “Other”. These 104 labeled set of time series TS serves as the time-series learning set. It can be noted that the learning set contains equal samples of “Earthquake events” and “other”. This is done to achieve a balanced training set to avoid classifier bias during the detection of earthquakes. In the absence of a balanced dataset, resampling techniques can be used to deal with class imbalance \cite{chawla2004editorial}. Using the steps mentioned in Section 2 in conjunction with Algorithm 1, 8 shapelets along with their respective IGs are found as shown in Fig 3. The highlighted section of the series is the shapelet and it occurs with the onset of an impulsive waveform and stops after the end of the event. The detection of earthquakes from continuous time histories can be treated as a binary classification problem as to whether a time series contains an earthquake event or not. Hence two families of shapelets have been discovered which is evident from Fig.3. EQShapelets 1,2,4 -7 are extracted from the time series belonging to the class “Earthquake events” and EQShapelets 3 and 8 are from class “Others”. The remaining 1716 time series in the one-week dataset is transformed using the discovered shapelets to construct a 1716 x 8 matrix containing the minimum distance between each shapelet and the time series.  Random Forest classifier with 500 trees is used on this shapelet transformed data to detect earthquake events from time history recorded between 9 January 2011 (00:00:00) and 15 January 2011 (00:00:00). Shapelets were able to detect almost 200 more events with lower false negatives than the FAST approach used in \cite{yoon2015earthquake}. Also, all catalog events (13/13) were correctly detected by the EQShapelet-based classifier as shown in Fig.4. (Note: The NCSN catalog has 24 cataloged events out of which 11 events occurred on 8 January 2011 and rest of the 13 events occurred between 9 – 14 January 2011). 

A prediction probability was also returned by the classifier for each of these detections. Out of the 299 events, 95 events were detected with a high prediction probability between 96\% and 100\% while 41 events buried in noise were detected with a probability of 50\% – 59\%. Nearly 46\% of the detected events have a prediction probability of 90\% and higher. The newly detected events are compared with signals on all three components (EHE, EHN, EHZ) of the data recorded at station CCOB. The waveforms are carefully inspected to make sure that the waveforms classified as true earthquake events resemble an impulsive earthquake signal on all three channels although shapelets were only trained on the EHN channel for detection. The noisy events detected with a probability between 50\% to 60\% are carefully checked for the presence of false positives and false negatives. There were 7 false negatives (missed detections) and 11 false positives (wrong detections). Therefore, the proportion of detected events that are true events i.e., precision is 96.3\% and recall is 97.6\%. The detection sensitivity of shapelets is clearly high which shows the robustness of the shapelet-based classifier for identifying earthquake events. A shape-based approach such as shapelets can solve a wide range of research problems in seismology pertaining to the shape of seismic waveforms. Shapelets, if implemented at a large scale, can significantly reduce catalog completeness magnitudes and can serve as an effective tool for earthquake monitoring and cataloging.  

\subsubsection{Identification of strong velocity pulses in ground motion}
Earthquake engineers are particularly interested in near-fault ground motions with impulse behavior that has the potential to impose extreme seismic loads on structures. Various techniques have been proposed in the literature \cite{ertuncay2019alternative,chang2016improved,shahi2014efficient,mavroeidis2003mathematical,mena2011selection,ghaffarzadeh2016classification,kardoutsou2017new} to identify such pulse-shape signals including mathematical constrains, wavelet transformations, cross correlation, and others. A shape-based approach such as shapelets can serve as a useful pulse identification tool to easily identify a velocity waveform as a pulse-shape or non-pulse-shape. The pulse identification capability of shapelets is tested on a ground motion dataset containing crustal earthquakes obtained from the NGA-West2 research project \cite{doi:10.1193/072113EQS209M}. 

\begin{figure}[htbp]
  \centering
  \captionsetup{justification=centering}
  \includegraphics[scale=0.70]{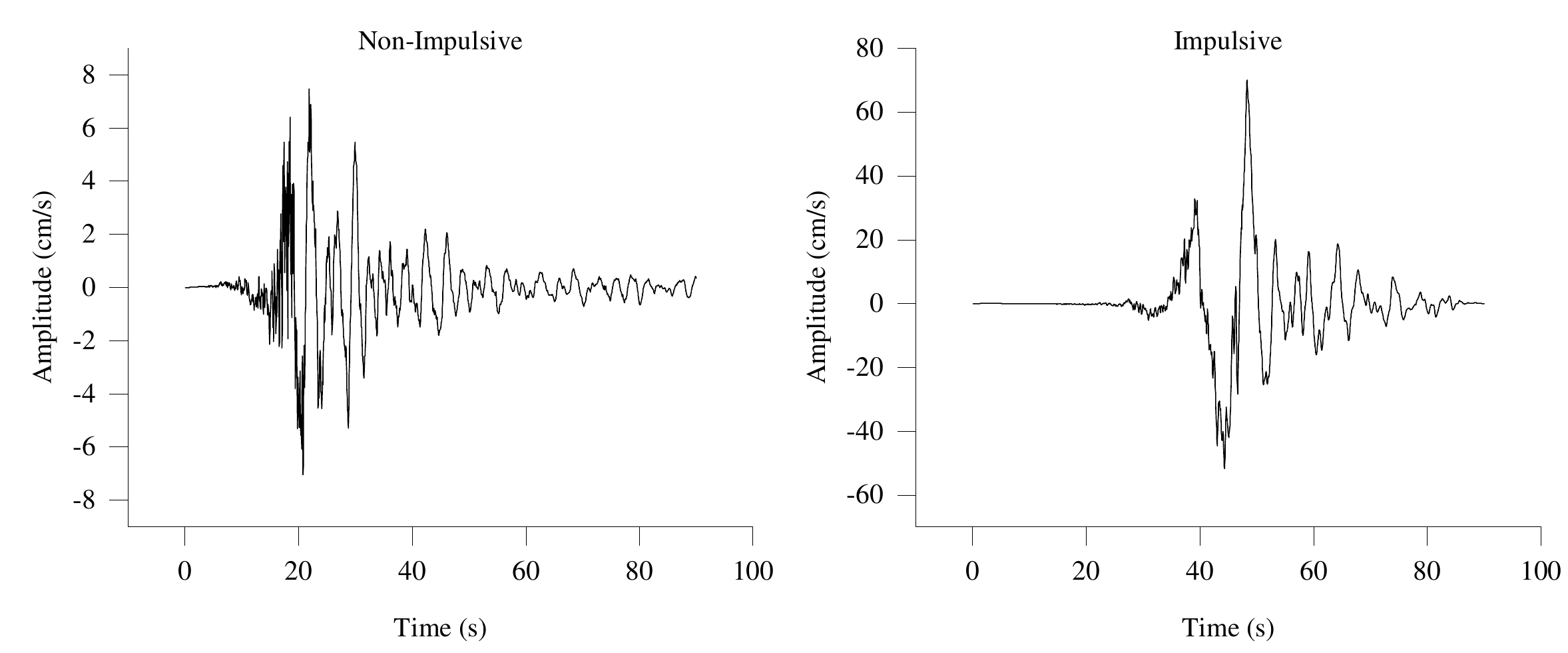}
  \caption{Example of non-impulsive vs impulsive velocity waveform}
  \label{fig:fig5}
\end{figure}

\begin{figure}[htbp]
  \centering
  \captionsetup{justification=centering}
  \includegraphics[scale=0.75]{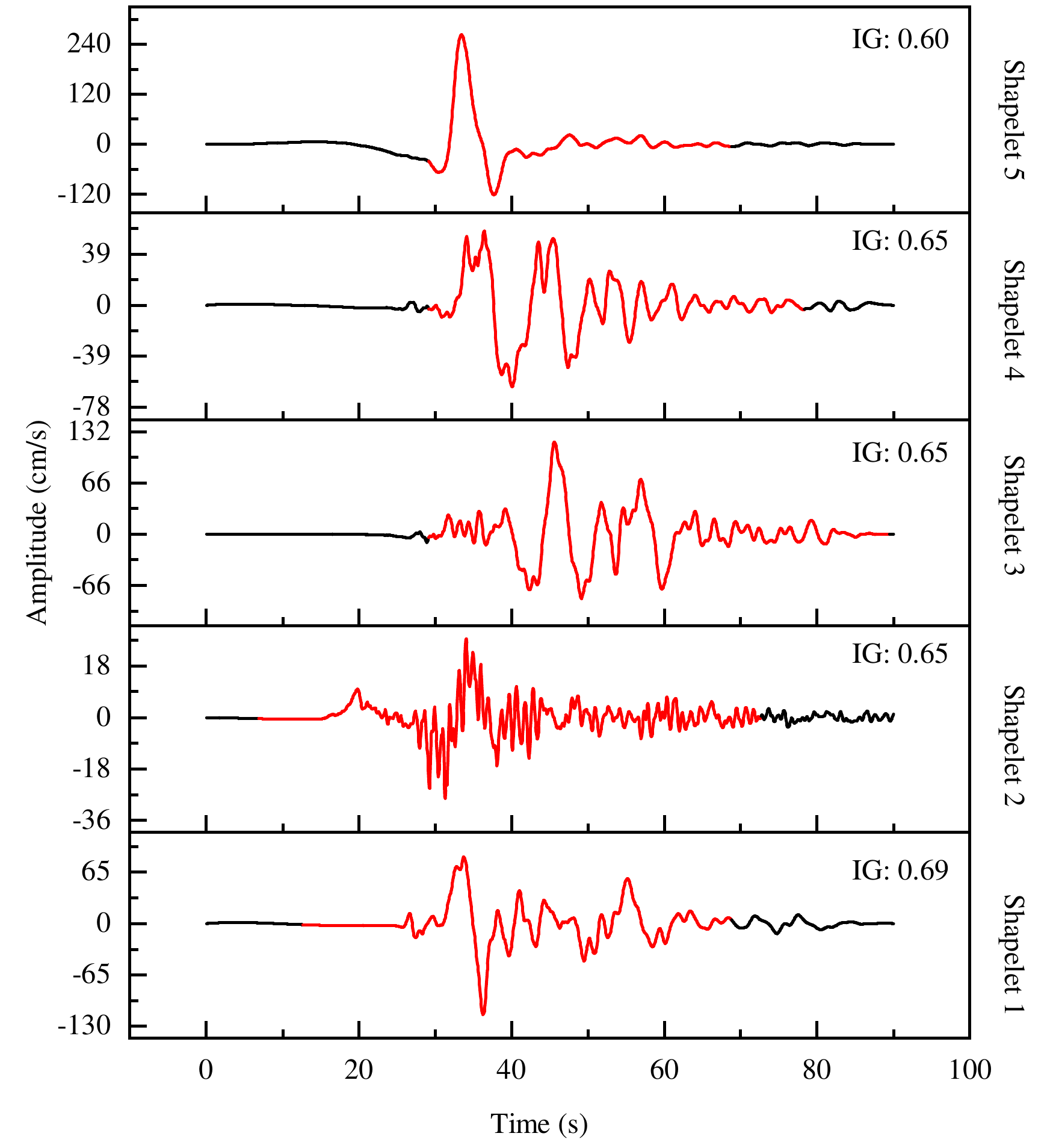}
  \caption{Examples of shapelets discoverer for the identification of strong velocity pulses in ground motion data}
  \label{fig:fig6}
\end{figure}

After denoising, the acceleration waveforms are integrated to obtain the velocity waveforms to easily visualize the pulse-shaped signals. In addition to visual investigation, criteria used in \cite{shahi2014efficient} is used to manually label the waveforms as “impulsive” or “non-impulsive”. Fig.5 shows an example of an impulsive and non-impulsive velocity waveforms. The impulsive waveforms have pulse-shaped signals with long and large amplitudes compared to the other waveform. A total of 100 waveforms of equal length are selected such that 50 are impulsive and the other 50 are non-impulsive waveforms. These 100 waveforms are split into training (60\%) and testing sets (40\%). The training set containing 60 time series is used to identify potential shapelets. A total of 98 shapelets are discovered using Algorithm 1 and the first 5 shapelets having the highest information gain are shown in Fig.6 for illustrative purposes. 

Shapelets 1 and 3-5 capture the pulse portion of the time series as the important shape criterion for classification. Shapelet 2 captures almost the entire non-impulsive time series as a shape. Since this is a binary classification problem, the shapelet algorithm selects distinct shapes for each of the two classes for accurate classification. Among the 98 discovered shapelets, 40 shapelets correspond to non-impulsive time series while the other 60 captures the various shapes of pulses present in the impulsive time series. The test set containing 40 time series (20 impulsive and 20 non-impulsive) is then transformed using the discovered shapelets to construct a 40 x 98 matrix containing the minimal distance between each shapelet and time series.  This serves as the input to a Random Forest classifier with 500 trees. The shapelet-based classifier correctly identified 18/20 impulsive waveforms and 16/20 non-impulsive waveforms. For 12 of the 18 detected impulsive waveforms, the classifier returned a prediction probability of above 90\%.  The rest of the events had a prediction probability between 75 – 89\%. The detection accuracy can be further increased by training the shapelet algorithm on a wide variety of pulse waveforms. It should be noted that in this experiment time series waveforms of equal lengths are used for shapelet finding. However, seismic waveforms of different lengths can be handled by using Dynamic Time Warping \cite{shah2016learning} as a similarity measure in the shapelet finding algorithm.

\subsection{Application of shapelets to wind and wind-induced vibration data}
\subsubsection{Identification of thunderstorms}
Identification of thunderstorms from continuous wind speed measurements is a topic of great interest to the engineering community to better understand the wind-induced response of structures and to study the distribution of extreme wind speeds \cite{solari2014emerging}. There are currently two categories of methods for the classification of thunderstorms: meteorological-based and wind engineering-based methods. The meteorological methods examine wind events through exploration and reconstruction of the meteorological conditions using anemometers, barometers, radar and satellite images that contain information about the main meteorological parameters \cite{craig1976vertical,wakimoto1982life,gunter2015high}. On the other hand, rather examining the detailed meteorological information of the wind event, the wind-engineering based methods focus more on developing statistical parameters from the extreme wind velocities for the separation and classification of thunderstorms and to study their effects on structures \cite{gomes1976thunderstorm,riera1989pilot,twisdale1992research,choi1999extreme,choi2002gust,kasperski2002new,cook2003extreme,lombardo2009automated}. These methods suffer from severe limitations as several statistical parameters are not easily available in which cases expert judgment is required to visually inspect every time series record. This makes some extreme wind event identification methods in the literature questionable as they only involve the use of statistical parameters. A time series shape-based approach such as shapelets has the potential to minimize the above-mentioned limitations as shapelets are dependent only on the shape of the time series and not on its statistical parameters. 

\begin{figure}[htbp]
  \centering
  \captionsetup{justification=centering}
  \includegraphics[scale=0.80]{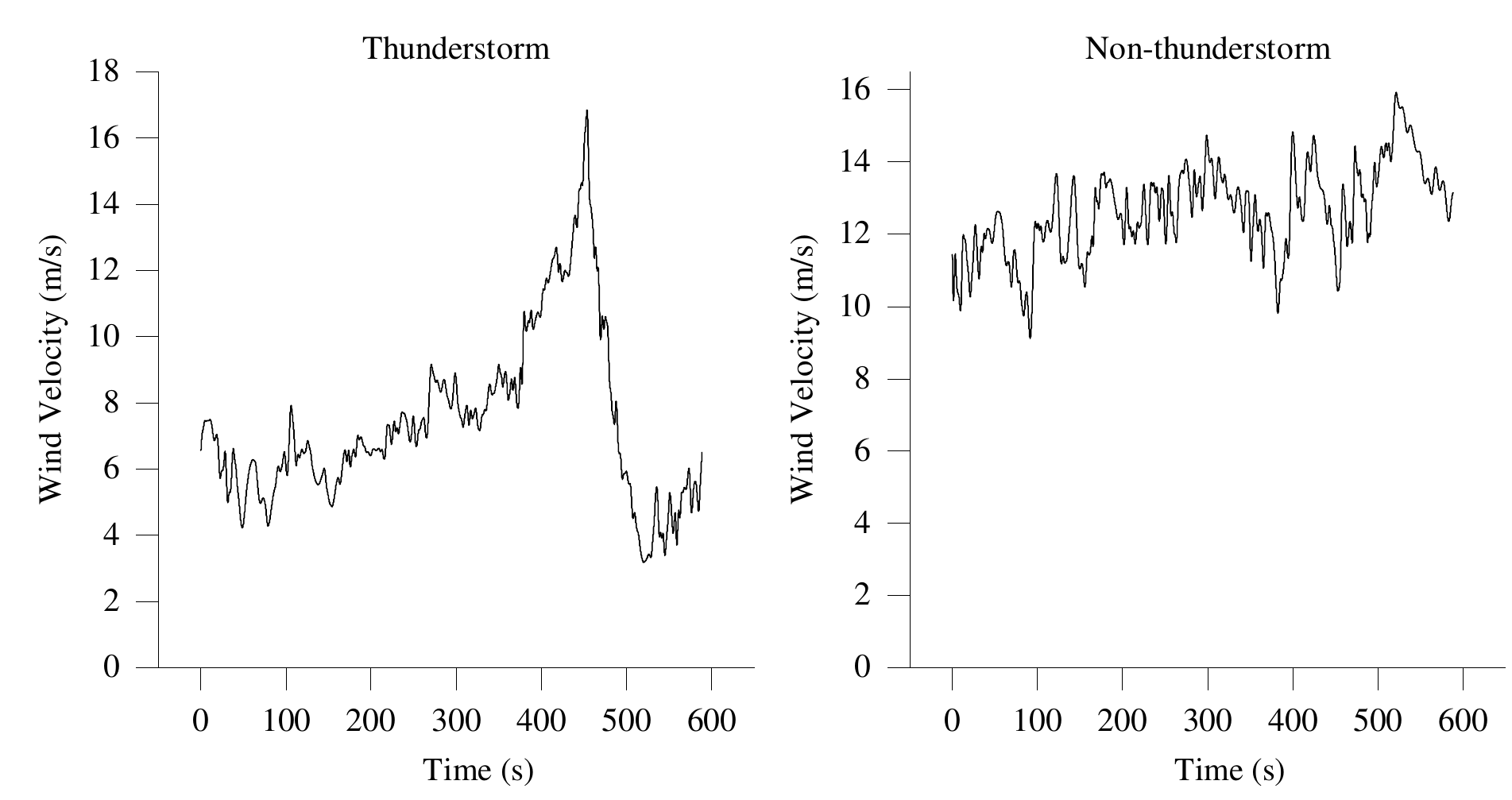}
  \caption{Time history of wind velocity measurements for a thunderstorm vs a non-thunderstorm}
  \label{fig:fig7}
\end{figure}

Fig. 7 shows the time history of wind velocity measurements from an anemometer during a thunderstorm versus a non-thunderstorm period. By visual inspection, thunderstorm time histories can be easily identified by the presence of a significant peak in wind velocity or by the absence of long stretches of aperiodic waveforms as seen in non-thunderstorm wind measurements. Shapelets can be employed to perform the same task by extracting the discriminative segments between the time series and comparing it with other time series to identify and extract the time series measurements corresponding to thunderstorms. The thunderstorm identification capacity of shapelets is tested on a dataset containing 30 time histories (15 thunderstorm records and 15 non-thunderstorm records) of wind speed measurements obtained from an anemometer. An equal number of thunderstorm and non-thunderstorm records are selected to achieve a balanced training set to avoid classifier bias and in the absence of a balanced dataset, resampling techniques can be used \cite{chawla2004editorial}. Each time history is 10-min long measured at a sampling frequency of 10 Hz. The dataset is preprocessed to remove noise and other spurious signals and is split into training (60\%) and testing sets (40\%) where shapelets are identified from the training set. 

\begin{figure}[htbp]
  \centering
  \captionsetup{justification=centering}
  \includegraphics[scale=0.70]{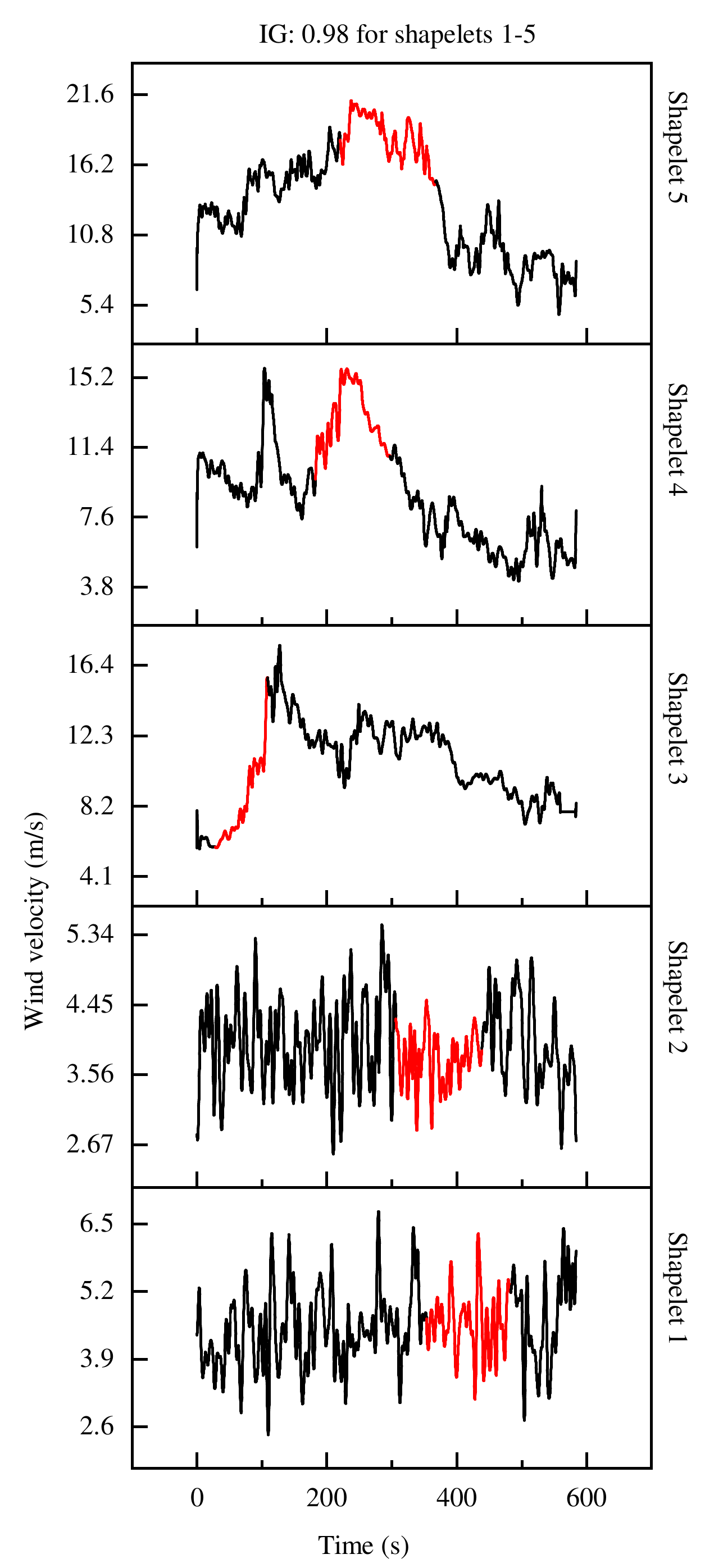}
  \caption{Examples of shapelets discovered for the separation and classification of thunderstorms}
  \label{fig:fig8}
\end{figure}

Using Algorithm 1, a total of 48 shapelets are discovered and the top 5 shapelets each with an Information Gain (IG) of 0.988 are shown in Fig. 8.  Since this is a binary classification problem (thunderstorm vs non-thunderstorm), two families of shapelets are discovered, one for each class. As anticipated, the shapelet algorithm extracts the peaks as the discriminative shapes as seen from shapelets 4 and 5. Besides, the algorithm also extracts the rise in wind velocity as a discriminatory shape as seen from shapelet 3. Shapelets 1 and 2 correspond to waveforms from the non-thunderstorm time series records that do not exhibit significant changes in the wind velocity. The testing set is then transformed using the 48 shapelets using Algorithm 2 to construct a 30 x 48 matrix where each element corresponds to the minimum Euclidean distance between each shapelet and time series. A shapelet-based Random Forest classifier with 500 trees is then used on the shapelet-transformed test set. A 100\% classification accuracy is obtained meaning the shapelet-based classifier was able to accurately identify and separate every thunderstorm time series from non-thunderstorm time series.  All the detected thunderstorm events were associated with a prediction probability of over 85\%. 

The key motivation behind this experiment is to show that shapelets provide a level of simplicity and transparency combined with accuracy compared to the other state-of-the-art approaches. Thus, shapelet-based classifiers can serve as an effective tool to extract and catalog thunderstorms from large-volumes of data produced by long-term wind monitoring networks. The applicability of shapelets, in this case, can be widened in two ways. Most wind monitoring networks employ anemometers, LIDARs and barometers to study extreme wind events. So along with wind velocity measurements, data about pressure, temperature, and change in wind direction are also recorded. Shapelet transforms for multivariate time series \cite{bostrom2018shapelet,bostrom2017shapelet} can be used in such cases to capture the multivariate features which lead to a better understanding of the phenomenon and better classification accuracy. Also, apart from thunderstorms, shapelets can be used to effectively identify other extreme wind events like depressions, gust fronts, and cyclones.

\subsubsection{Identification of Vortex-Induced Vibrations (VIV) in bridges}
Vortex-induced vibrations (VIV) are a particular vulnerability of long-span bridges \cite{larsen2000storebaelt,frandsen2001simultaneous,fujino2002wind,li2011investigation}. Owing to their slender design and flexibility, these structures undergo large-amplitude vibrations even under low ambient winds thus increasing the occurrence probability of VIVs. Although this may not be an apparent structural safety issue, the resultant fatigue can lead to failure of components and may also cause discomfort to drivers. To assess the wind-induced dynamic performance of these structures, numerous long-term structural health monitoring programs have been installed on long-span bridges. They serve as a means to periodically check the status of the structure and stave off any developing issues.  However, this continuous monitoring generates a large volume of data, which contains salient information about bridge behavior.  However, VIV constitutes only a small portion of the measured data as the majority of the structural response recorded by the sensors is due to random excitations like buffeting and vehicle-induced vibrations. In many studies, VIV was identified by visual inspection of measured data. A few other methods used machine learning techniques for automatic identification of VIV \cite{li2018data,li2017cluster,xu2019prediction}. These methods were not able to achieve high accuracy for identifying VIV as the input features were based on spectral representation of the data and the machine learning parameters were based on user discretion.

Shapelets can be advantageous in this case to intuitively separate VIV from other vibrations. Fig.9 shows the acceleration time history of a long span bridge experiencing buffeting and vehicle-induced vibrations while Fig. 10 shows the acceleration time history of the same bridge under vortex-induced vibration. Based on visual inspection, it is easy to differentiate between the two vibrations. However, it is quite difficult to use the raw time histories for shapelet extraction due to the long periods of sinusoidal waveforms present in both the vibrations. This can be overcome by extracting the envelopes of the acceleration time history which gives an overall shape to the vibration time series. The envelopes can then be easily used as input for the discovery of shapelets. Fig. 9 and Fig. 10 show the upper and lower root-mean-square (RMS) envelopes of the bridge acceleration time history calculated using a moving window.

\begin{figure}[htbp]
  \centering
  \captionsetup{justification=centering}
  \includegraphics[scale=0.85]{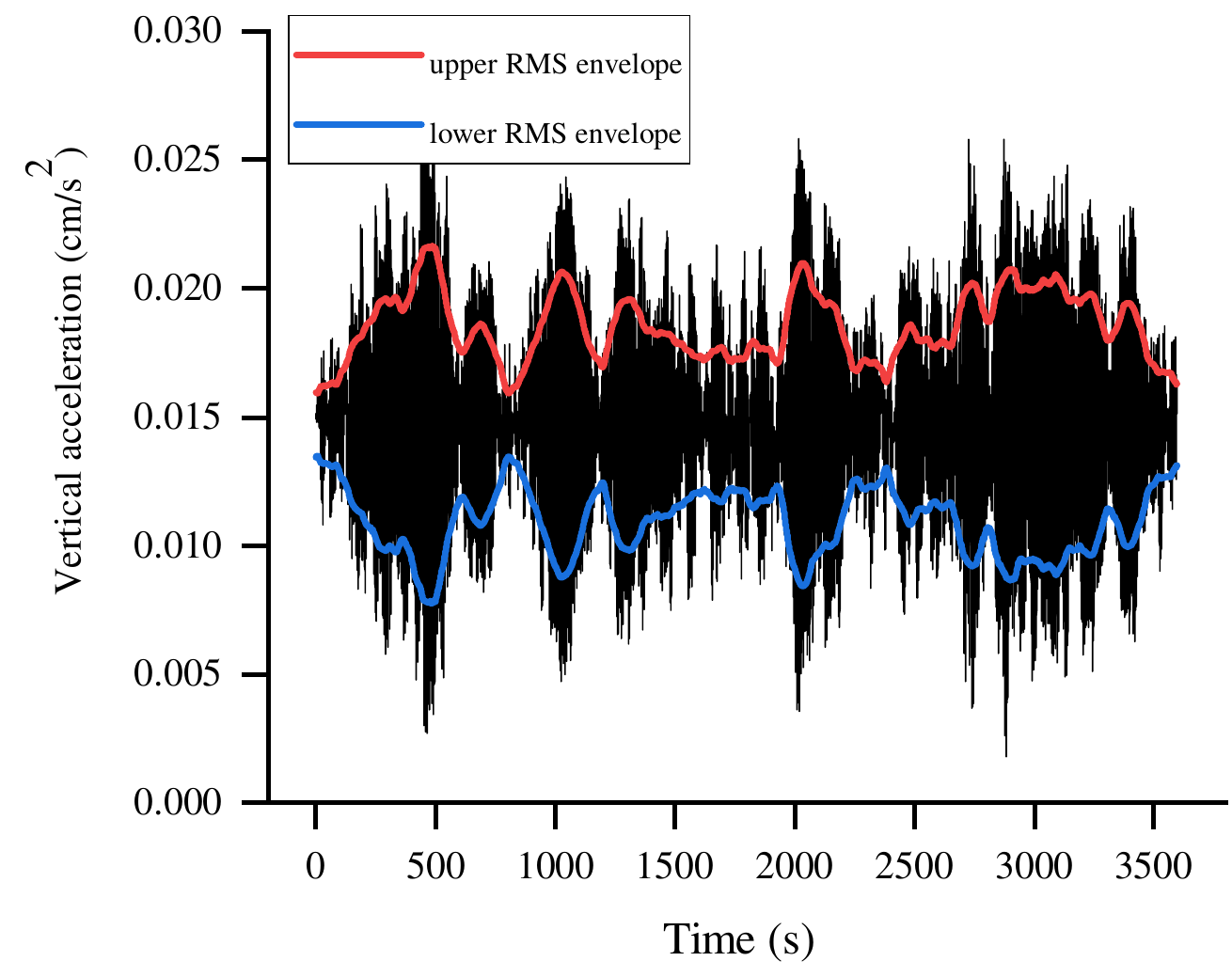}
  \caption{Acceleration time history of a bridge}
  \label{fig:fig9}
\end{figure}

\begin{figure}[htbp]
  \centering
  \captionsetup{justification=centering}
  \includegraphics[scale=0.85]{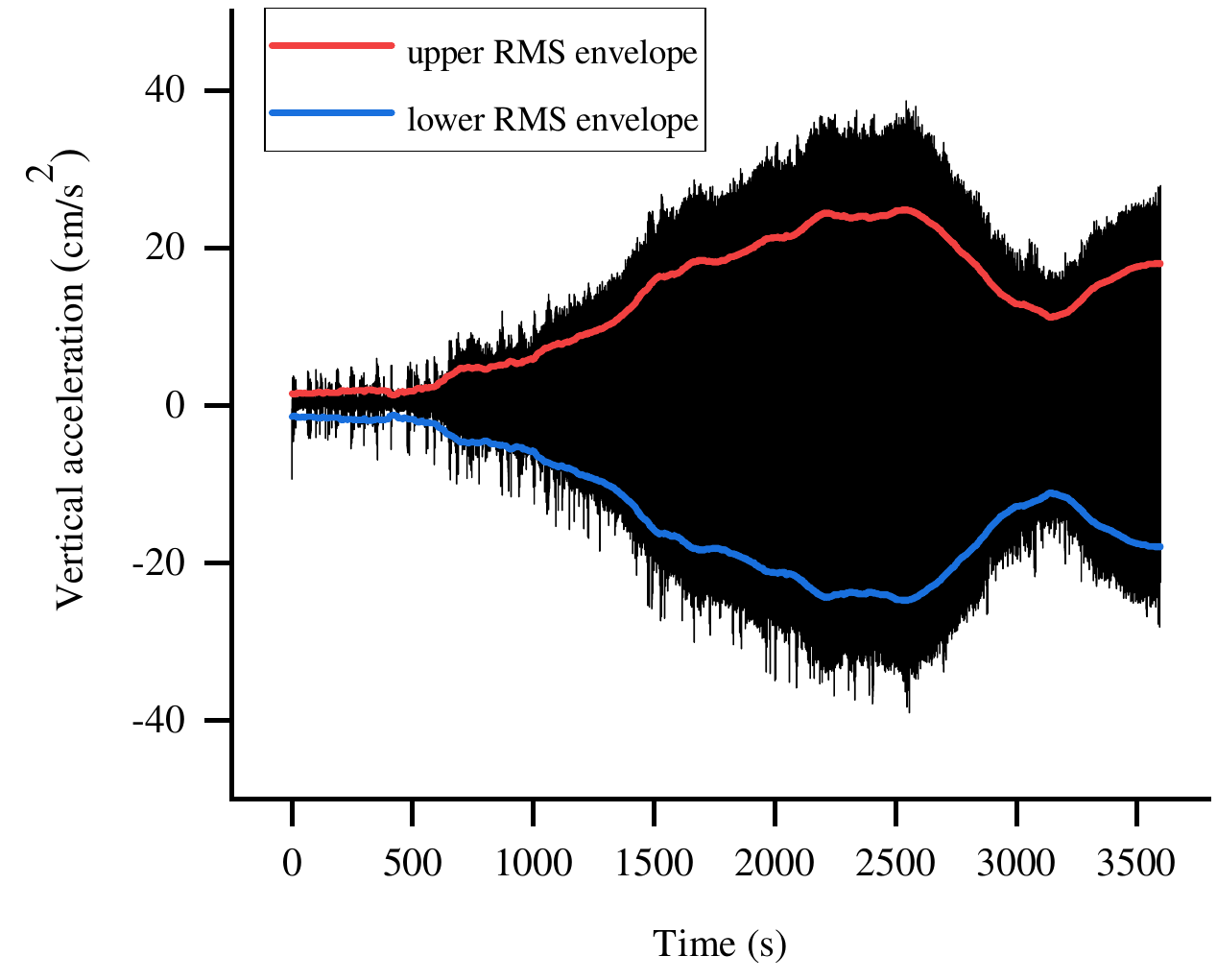}
  \caption{Acceleration time history of a bridge experiencing Vortex-Induced Vibration (VIV)}
  \label{fig:fig10}
\end{figure}

RMS envelope is used here instead of a peak envelope, as RMS provides a better insight into the amplitude of the signal even when noisy or spurious signals are present. By looking at the RMS envelopes of both the vibrations, the separation of VIV has become a much easier task now. The envelope of the acceleration time series in Fig.7 shows multiple sharp peaks when compared to the envelope of the VIV time history. The detection capacity of the shapelets is tested on a dataset containing 25 VIV vibrations time histories and 25 time histories of other vibrations. The upper RMS envelope of each of these time histories is extracted as it is sufficient to use one of the two envelopes for shapelet extraction. The time histories are then manually labeled as “VIV” or “Other “to form the time series learning set. The dataset is split into training (60\%) and test (40\%) sets where the training set is used to identify the shapelets.

A total of 36 shapelets were discovered using Algorithm 1 and the top 5 shapelets each with an Information Gain (IG) of 0.99 are shown in Fig.11. As expected, the shapelet algorithm extracts the multiple peaks in the envelope of time history that corresponds to buffeting and vehicle-induced vibration. The shapelets are then used to transform the test data and a random forest classifier with 500 trees is used for classification. Classification accuracy of 95\% was achieved as the shapelet-based classifier was able to correctly classify 19/20 test datasets where the prediction probability associated with each of the correctly classified events was above 90\%. The simple experiment serves as an example of how an intuitive time series shape-based approach like shapelets can serve as an effective tool for autonomous structural response classification.

\begin{figure}[htbp]
  \centering
  \captionsetup{justification=centering}
  \includegraphics[scale=0.80]{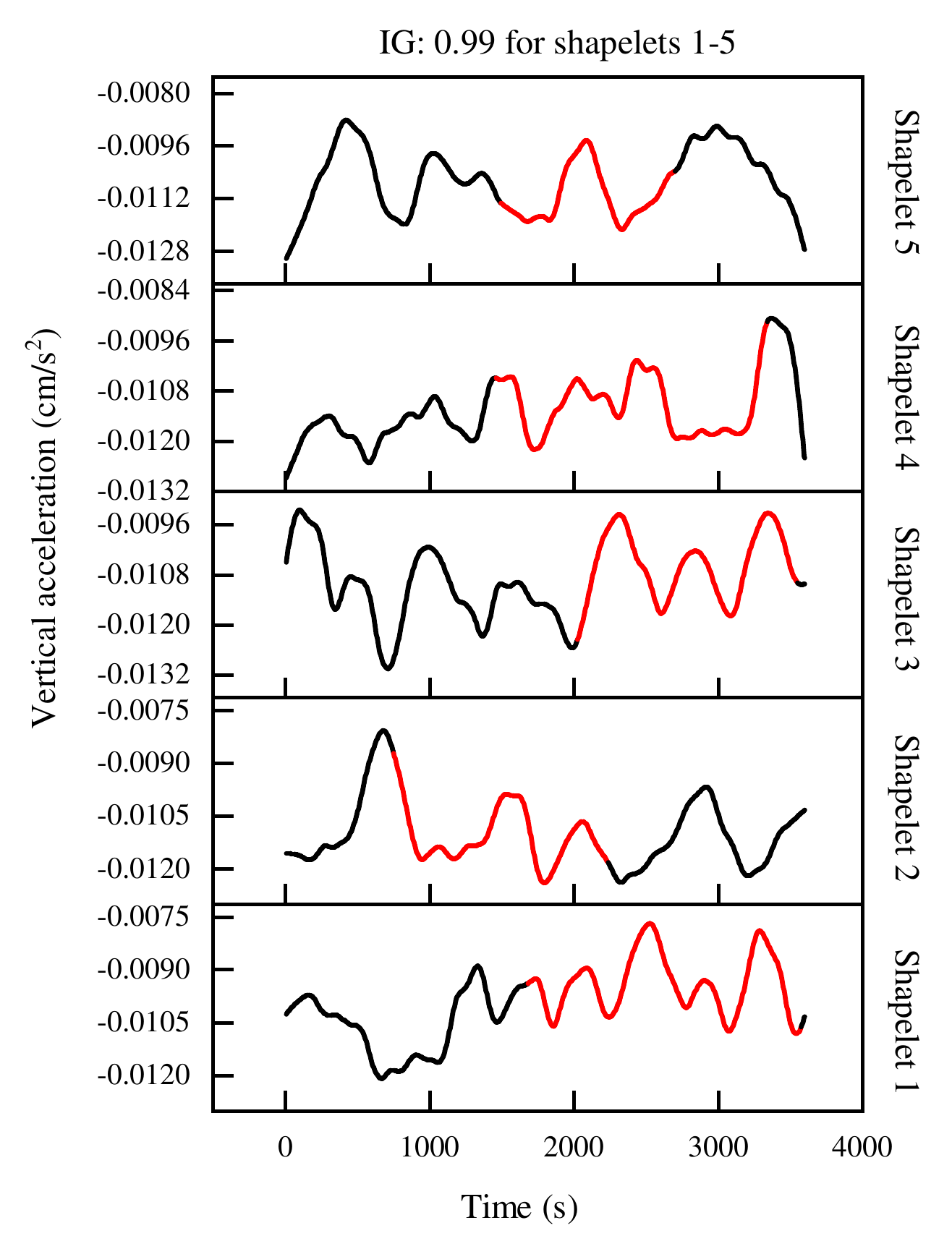}
  \caption{Examples of shapelets discovered for the identification of VIV events}
  \label{fig:fig11}
\end{figure}

\subsubsection{Other applications of shapelets in wind engineering}
Wind-induced motion is a particular vulnerability of tall buildings and in recent years the civil engineering community has become increasingly committed to understanding the behavior of tall buildings post-construction using long-term structural health monitoring programs. These programs generate a large amount of structural response data in terms of building acceleration. Shapelets can be used to classify the response of the structure to a variety of environmental conditions such as thunderstorms, sandstorms, hailstorms, and other sustained wind events. An envelope analysis (as used in the identification of VIV in bridges) can be used on the structural acceleration data to aid shapelets in structural response classification.

\subsection{Application of shapelets to wave data}
\subsubsection{Detection of breaking waves}
Among the various loads on offshore structures, the high impulsive slamming forces due to plunging breaking waves are one of the most significant types of loads. Usually, theoretical and experimental studies are used to identify the occurrence of plunging breakers \cite{robertson2013breaking,perlin2013breaking,liu2011new}. In practice, these criteria are limited to specific bathymetry, water depth, and other conditions, which makes it difficult to apply these criteria in general. For example, the most common method to detect plunging breakers is using surf similarity parameters in which the seafloor slope is an important parameter. This criterion is therefore not suitable for the detection of slamming events at a structure that is located on a flat sea bottom. Using an unsuitable plunging criterion or only breaking criterion may lead to an incorrect detection result.  Since wave tests in the laboratory usually provide a large number of data, machine learning, in particular, neural networks have been used as an alternative approach in the literature \cite{deo2003prediction,akoz2011prediction,kouvaras2018machine,tu2018detection} to address the problem of plunging breaking wave detection at a specific site. Most of these studies using machine learning techniques were trained on a limited number of datasets that were produced under certain laboratory conditions. This limits the applicability of these methods in predicting the characteristics of braking waves outside the range of the training dataset.

Shapelets can serve as an effective tool, in this case, to detect and identify plunging breaking waves. The detection capability of shapelets is tested on the experimental data from the WaveSlam project \cite{arntsen2013data}. The project aims to study the slamming forces from plunging breaking waves on jacket structures. More details about the experiment can be found at \cite{FZK2013W30:online,taprojec97:online}. The wave elevation data from Wave Gauge S9 (WGS9) is used for the present study. For each test, whether the regular waves are breaking or not is identified based on visual observations and labeled in the dataset. The sampling frequency of wave gauges is 200 Hz. A representative time history of measured wave elevation data is shown in Fig.12.  The measured wave elevation data is first pre-processed using a band-pass filter to remove high frequency noise. During each run, 20 waves were recorded. Zero up-crossings are used to extract individual waves from the measured data. Fig. 13 shows a plunging breaking wave vs a non-breaking wave extracted from the wave elevation data. It can be clearly seen that the geometrical properties of the two waves are quite different. The breaking wave is asymmetrical and has a sharp crest and deep trough when compared to the non-breaking wave. From the experimental data, 100 waves (along with their labels) are selected at random such that 50 are plunging breaking waves and 50 are non-breaking waves. This constitutes the time series learning set for shapelet transform. The dataset is split into training set (60\%) and test set (40\%) where shapelets are identified using the training set.

\begin{figure}[htbp]
  \centering
  \captionsetup{justification=centering}
  \includegraphics[scale=0.65]{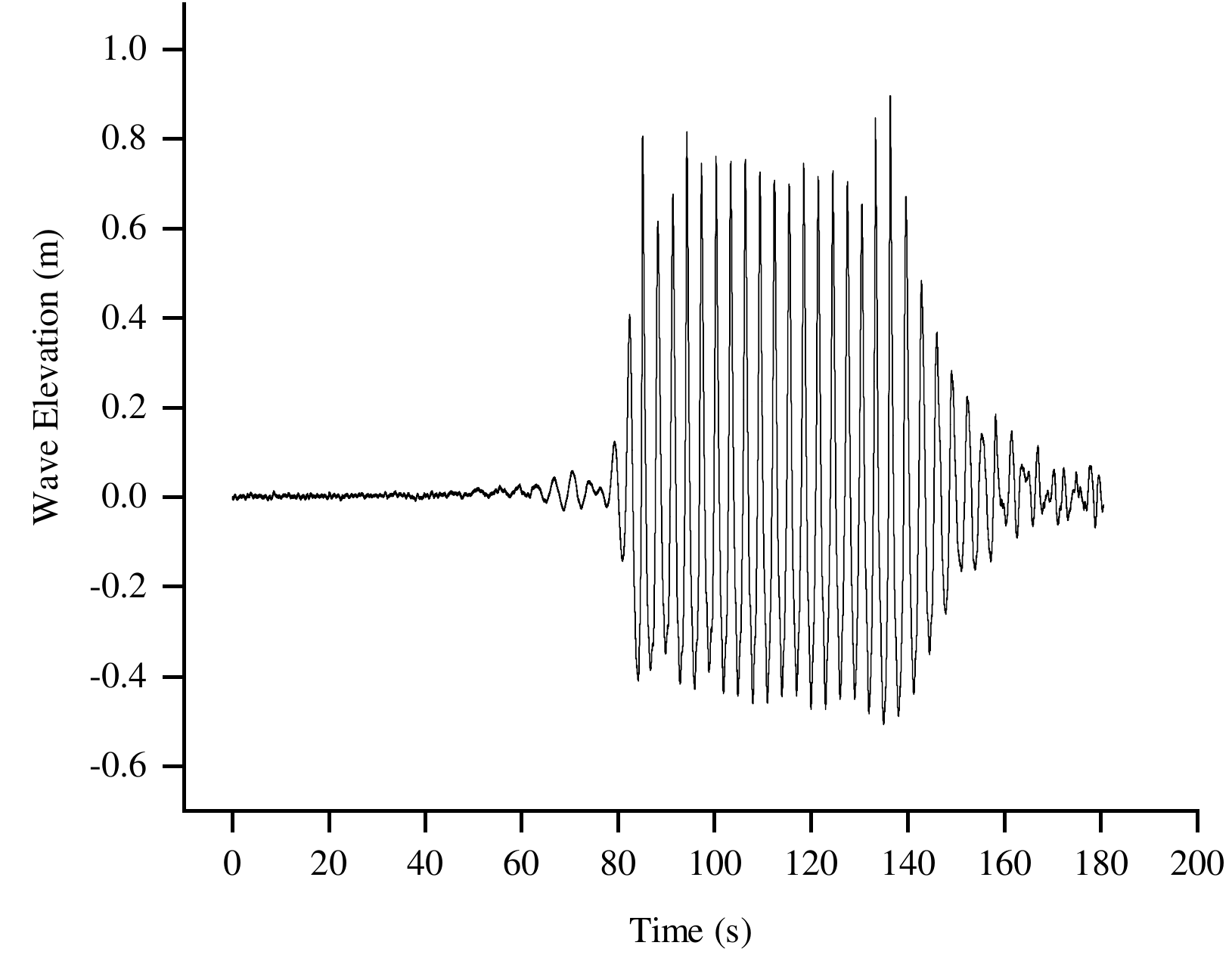}
  \caption{A representative time history of measured wave elevation during a wave test}
  \label{fig:fig12}
\end{figure}

\begin{figure}[htbp]
  \centering
  \captionsetup{justification=centering}
  \includegraphics[scale=0.65]{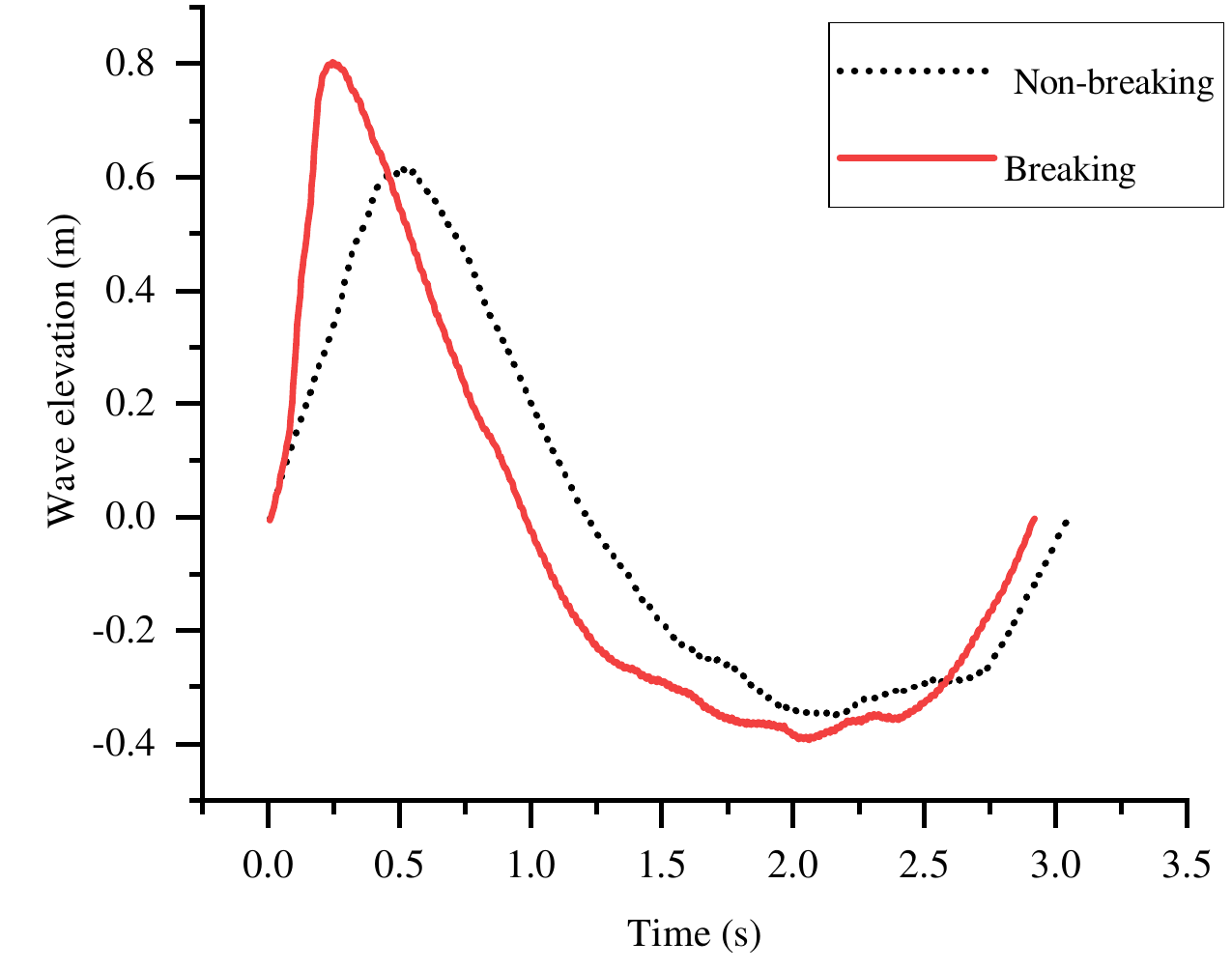}
  \caption{A non-breaking wave and a breaking wave extracted from the wave elevation time series}
  \label{fig:fig13}
\end{figure}

\begin{figure}[htbp]
  \centering
  \captionsetup{justification=centering}
  \includegraphics[scale=0.75]{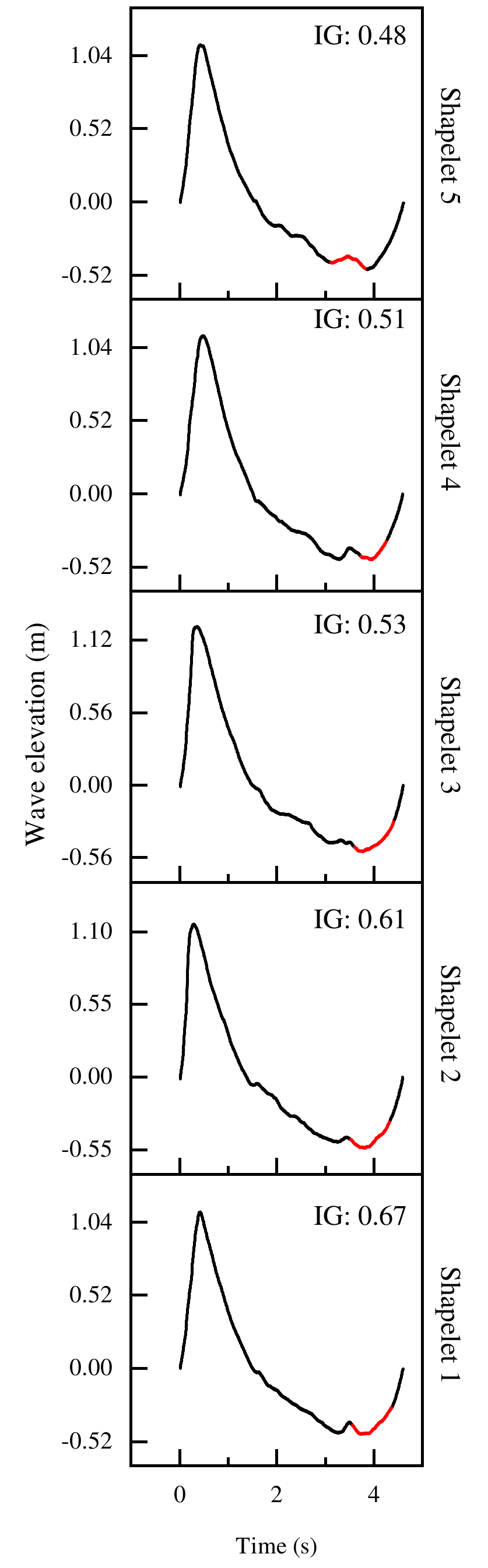}
  \caption{Examples of shapelets discovered for the detection of breaking waves}
  \label{fig:fig14}
\end{figure}

Using the shapelet algorithm, a total of 47 shapelets are discovered. The top 5 shapelets along with their respective IGs are shown in Fig. 14.  The algorithm surprisingly identifies troughs of waves as the discriminative shapelets as opposed to the crest. As mentioned earlier, during each run of the laboratory experiment, 20 continuous breaking waves or non-breaking waves were produced. So, the trough of the wave (as seen in Fig.11) extracted using the zero up-crossing method might serve as a precursor for the breaking onset of the next wave. This is well in agreement with the observations in the literature that the depth of the trough in front of the breaking crest reaches its maximum close to breaking. From the discovered shapelets it can be concluded that even if two waves have very similar geometric properties, troughs can be used as an effective indicator to identify a breaking wave. The breaking wave classification problem can now also be used as a prediction problem where the troughs can be used to predict if the next wave will break or not. The extracted shapelets are used in transforming the test data and a shapelet-based Random Forest classifier with 500 trees is used on the shapelet-transformed test set. 97\% classification accuracy is obtained meaning the shapelet-based classifier was able to accurately classify 39 of the 40 waves in the test set. Out of the 39 events, the classifier returned a prediction probability of over 90\% for 23 events, over 85\% for 7 events and over 80\% for 9 events. This experiment demonstrates the effectiveness of using an intuitive shape-based approach like shapelets that provide accuracy combined with interpretability that can help domain experts better understand their data, like in the present case of identifying breaking waves.

\subsubsection{Other applications of shapelets to wave data}
Structural damage detection in offshore platforms has been of great interest to researchers as these structures are highly vulnerable to damage from extreme environmental conditions. As a result, several structural health monitoring programs have been installed on offshore structures that provide continuous data about the performance of the structure. Shapelets can be used on the data from the sensors to identify and classify time histories corresponding to structural damage as the shape of the response time history of a damaged structure shows significant differences from that of a healthy structure. Also, erroneous data detection from sensors in the ocean buoys is a potential application of shapelets. Shapelets can be used as automatic quality control to identify and separate corrupt data/instrument errors from real data.

\section{Concluding remarks}
In this paper, a new time-series representation named Shapelet transform is proposed to detect and extract understandable and easily visualizable shape-based features from time series data. Shapelet transform provides a universal standard feature, based on the distance between a shapelet and a time series, for all time series mining tasks. This way, shapelet transform can help to autonomously detect events of interest from large databases generated as a result of long-term monitoring. The shapelet transform bundled with a Random Forest classifier used in the present study, serves as a “white-box” machine learning model with understandable features and transparent algorithm thus making the detection results fully explainable. In this paper, shapelets are used to identify characteristic shapes of different time series in earthquake, wave and wind data. The extracted shapes are then used as discriminative segments to convert the new incoming time series into a local-shape space called shapelet transform. The shapelet transform in combination with a Random Forest classifier with 500 trees is used to identify earthquake events from continuous seismic data, identify strong velocity pulses from ground motion data, identify thunderstorm events from wind speed measurements from anemometers, identify large-amplitude vortex-induced vibration of bridges and to detect the occurrence of plunging breaking waves. Other potential applications of shapelets to earthquake, wind and wave data have also been briefly discussed in this paper.

\section*{Data and resources}
Continuous waveform data and earthquake catalogs for this study were last accessed in August 2019 through the Northern California Earthquake Data Center (NCEDC), doi:10.7932/NCEDC (Northern California Earthquake Data Center), operated by the UC Berkeley Seismological Laboratory and the U.S. Geological Survey (USGS). The ground motion dataset containing crustal earthquakes were obtained from the NGA West2 database in the PEER Ground Motion Database. The wind speed measurements of thunderstorms used in this research were recorded by the wind monitoring network set up for the European Projects “Winds and Ports” (grant no. B87E09000000007) and “Wind, Ports and Sea” (grant no. B82F13000100005), funded by European Territorial Cooperation Objective, Cross-border program Italy-France Maritime 2007–2013. The thunderstorm dataset is part of the project THUNDERR - Detection, simulation, modeling and loading of thunderstorm outflows to design wind-safer and cost-efficient structures – funded by European Research Council (ERC) under the European Union’s Horizon 2020 research and innovation program (grant agreement no. 741273) The thunderstorm and bridge datasets are proprietary and have some confidentiality restrictions. Access to WaveSlam data was provided by Prof. Øivind A. Arntsen, the WaveSlam project manager at the Norwegian University of Science and Technology (NTNU). Requests for such access may be directed to him. The basic algorithm for shapelet discovery is available at "Anthony Bagnall, Jason Lines, William Vickers, and Eamonn Keogh, The UEA \& UCR Time Series Classification Repository” (\url{www.timeseriesclassification.com}). Additional information related to this paper may be requested from the authors.

\section*{Funding}
This work was supported in part by the Robert M. Moran Professorship and National Science Foundation Grant (CMMI 1612843).

\section*{Acknowledgement}
We would like to thank Prof. Giovanni Solari and Prof. Massimiliano Burlando from the University of Genoa for sharing the thunderstorm datasets, Prof. Øivind A. Arntsen, from the Norwegian University of Science and Technology for providing access to the WaveSlam datasets, Dr. Anthony Bagnall from the University of East Anglia for providing access to the shapelet codes and Dr. Andrew Kennedy from the University of Notre Dame for providing his insights regarding the application of shapelets for ocean waves. We would also like to express our sincere gratitude to Mr. Deniz Ertuncay from the University of Trieste, Italy for directing us to the NGA-West2 database.

\bibliographystyle{unsrt}  
\bibliography{references}  





\end{document}